\pdfoutput=1

\documentclass[11pt, a4paper]{article}

\usepackage[acceptedWithA]{tacl2018v2}

\usepackage{times}
\usepackage{latexsym}

\usepackage[T1]{fontenc}

\usepackage[utf8]{inputenc}

\usepackage{microtype}
\usepackage{url}
\usepackage{amsmath}
\usepackage{amsthm}
\usepackage[inline]{enumitem}
\usepackage{multirow}
\usepackage{subfig}
\usepackage{algorithm}
\usepackage{algpseudocode}
\usepackage{xcolor}
\usepackage{inconsolata}
\usepackage{xspace}
\usepackage{soul}
\usepackage{float}
\usepackage{fancyvrb}
\usepackage{listings}
\usepackage{booktabs}
\usepackage{graphicx}
\usepackage{amsfonts}
\newfloat{lstfloat}{htbp}{lop}
\floatname{lstfloat}{Prompt}

\newfloat{lstdoc}{htbp}{lop}
\floatname{lstdoc}{Guideline}

\title{Interactive Topic Models with Optimal Transport}

\author{Garima Dhanania\textsuperscript{*}\quad Sheshera Mysore\textsuperscript{*}\textsuperscript{\textdagger}\thanks{*~Equal contribution.}\thanks{\textdagger~Currently a Postdoctoral Researcher at Microsoft.}\quad Chau Minh Pham \\\quad \textbf{Mohit Iyyer \quad Hamed Zamani \quad Andrew McCallum}\\
University of Massachusetts Amherst, \textsc{\small MA, USA}\\
\texttt{\{smysore,zamani,mccallum\}@cs.umass.com}\quad\texttt{\{gdhanania,ctpham,miyyer\}@umass.edu}}

\renewcommand\footnotemark{}
\begin{document}
\newcommand{\etmmethod}{\textsc{EdTM}\xspace} 
\newcommand{\etmmethodbe}{$\textsc{EdTM}_{\text{BE}}$\xspace} 
\newcommand{\etmmethodce}{$\textsc{EdTM}_{\text{CE}}$\xspace} 
\newcommand*\cirnum[1]{\raisebox{.5pt}{\textcircled{\raisebox{-.9pt} {#1}}}}
\newcommand{\etmscoringf}{$f_{\text{sim}}$\xspace}
\maketitle

\begin{abstract}
Topic models are widely used to analyze document collections. While they are valuable for discovering latent topics in a corpus when analysts are unfamiliar with the corpus, analysts also commonly start with an understanding of the content present in a corpus. This may be through categories obtained from an initial pass over the corpus or a desire to analyze the corpus through a predefined set of categories derived from a high level theoretical framework (e.g. political ideology). In these scenarios analysts desire a topic modeling approach which incorporates their understanding of the corpus while supporting various forms of interaction with the model. In this work, we present \etmmethod, as an approach for \emph{label name supervised topic modeling}. \etmmethod models topic modeling as an assignment problem while leveraging LM/LLM based document-topic affinities and using optimal transport for making globally coherent topic-assignments. In experiments, we show the efficacy of our framework compared to few-shot LLM classifiers, and topic models based on clustering and LDA. Further, we show \etmmethod's ability to incorporate various forms of analyst feedback and while remaining robust to noisy analyst inputs.
\end{abstract}
\section{Introduction}
\label{sec-etm-intro}
Topic models have had a long history of development and use for exploring document collections and representing documents and collections for downstream tasks \cite{jbg2017tmbook}. While topic models are most commonly associated with probabilistic generative models \cite{blei2003latent}, other approaches leveraging matrix factorization \cite{lund2019fine} and clustering embeddings from pre-trained language models \cite{thompson2020topic} have also been explored for topic modeling. Across these models, a valuable feature for practitioners is their ability to represent latent topics with interpretable descriptors such as word, sentence, or documents assigned to latent topics. 

While latent topics are valuable for several forms of analysis \cite{roberts2013structural, hoyle-etal-2019-unsupervised}, topic models fall short of practitioners' expectations when analysts don't wish to be biased by machine generated topics, wish to have topics capture their own understanding of the corpus, or analyze a corpus in a top-down manner through categories defined by an analytical framework \cite{hong2022scholastic, hjalmar2022compgt, jasim2021communitypulse}. One approach to address such concerns may involve labeling a training set of documents with categories, followed by classification or supervised clustering to label the whole collection \cite{finley2005supcl}. However, these prove to be time consuming due to the need to label data and brittle when the set of categories evolve as practitioners' understanding of the corpus changes.

An alternative explored in a large body of work on probabilistic topic models involves supervising latent topics. This work has explored use of document metadata \cite{card2018neural}, seed words per topic \cite{jagarlamudi2012incorporating}, and constraints to supervise and refine latent topics \cite{hu2011interactive}. However, generative topic models prove challenging to scale to large corpora \cite{lund2017tandemanchor}, are influenced by document lengths \cite{hong2010twittertm}, and are limited in their ability to leverage performant pre-trained language models \cite{hoyle2022neural}. While an emerging body of work has effectively leveraged pre-trained language models for unsupervised topic modeling \cite{pham2023topicgpt, wang2023goalex, thompson2020topic} exploration of contemporary pre-trained models for interactive topic modeling has been limited. Our work fills this gap.

\begin{figure}[t]
     \centering
     {\includegraphics[width=0.49\textwidth]{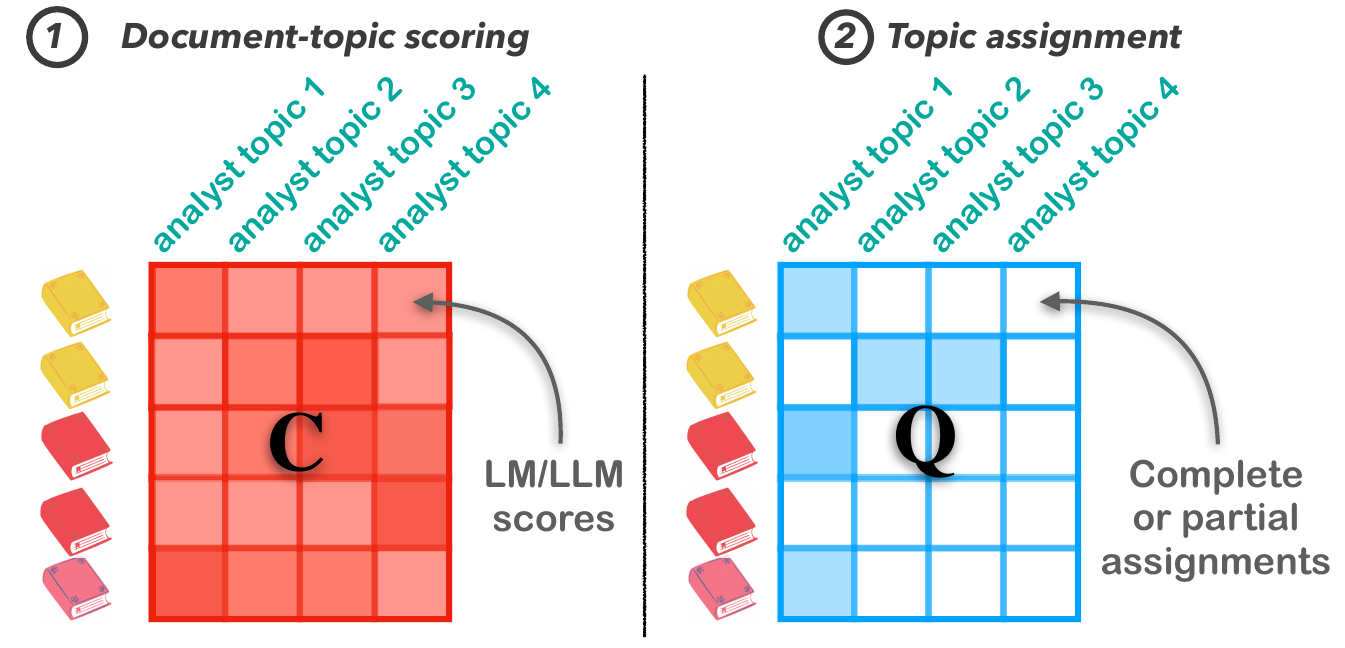}}
     \caption{Interactive topic modeling with \etmmethod consists of two steps, document-topic scoring for analyst provided topic names using LM/LLM bi-encoders and cross-encoders followed by computation of partial or complete topic assignments using optimal transport. Analyst topic names may take on various forms such as label names, descriptions, or documents, to support rich forms of interaction.}
     \label{fig-high-level}
 \end{figure}

In this work, we propose a framework for \emph{label name supervised topic modeling}. We show label names as a flexible form of interaction with topic models -- allowing analysts to specify topics as short label names akin to class labels, longer label descriptions, as well as seed documents. We model topic modeling as an \emph{assignment problem}, requiring a globally coherent assignment of documents to topics, and leverage the well explored machinery of optimal transport \cite{peyre2019computational} for interactive topic modeling. Use of optimal transport offers several advantages: assignment algorithms capable of leveraging GPU computation, an ability to leverage document-topic similarities obtained from pre-trained language models, and a mature body of work on assignment algorithms which may be applied for various topic modeling applications. We refer to our approach to interactive topic modeling as an \emph{editable topic model}, \etmmethod (Figure \ref{fig-high-level}). In experiments, we show \etmmethod to induce high quality topics compared to a range of baselines, support various forms of interaction from analysts, and induce robust topics even when presented with noisy analyst supervision. Code and datasets for our work will be released upon acceptance.
\section{Problem Definition}
\label{sec-etm-problem}
We consider a \emph{label supervised topic modeling} problem where we are presented with a collection of documents $D$ that an analyst is interested in studying and a set of topic labels $l \in L$ that the analyst would like to organize $D$ by or believes is contained in $D$. We are interested in predicting a partitioning of the documents in $D$ into the labels $L$ where each document could be assigned to multiple topics in $L$ or no topics at all. The latter case represents a selective topic assignment intended to capture the scenario where analysts intuitions of the topics $L$ are incorrect or incomplete, requiring that some documents or topics remain unassigned. We assume documents $d \in D$ to be short sentences or multi-sentence paragraphs.

Further, we are interested in supporting different forms of interactive analysis and exploration of $D$ - this is done through different forms of topic labels $l$. Topic labels may be represented as natural language \emph{names} for topics, example \emph{words or longer descriptions} of topics, or example \emph{documents} representing a topic. The various forms of topic labels combined with selective topic assignment allows for rich ways of interaction with a topic model. 

\section{Proposed Approach}
\label{sec-etm-method}
We propose a flexible framework for interactive topic modeling, \etmmethod, consisting of two components: a document-topic scoring leveraging expressive LM/LLMs, and document-topic assignment leveraging optimal transport assignment algorithms. We explore scores computed using pretrained LM bi-encoder models as well as expressive LLM crossencoders allowing analysts to specify their understanding of topics in a variety of intuitive natural language forms. For document-topic assignment, optimal transport algorithms allow assignment decisions to be made in a globally coherent manner and naturally incorporate document-topic affinities from expressive LLM scoring functions. Further, we leverage partial assignment algorithms \cite{benamou2015partialot} that selectively exclude high cost document-topic assignments, to ensure that topic-assignments remain robust to potentially noisy or incomplete labels provided by users -- a likely occurrence in interactive topic modeling. Finally, since computation of the $D\times L$ assignment remain expensive for large corpora with optimal transport, we compute approximate assignments through batched computation of assignments \cite{fatras2020batchedot}. We discuss the assignment algorithms and scoring functions used in \etmmethod next.

\subsection{Optimal Transport Topic Assignment}
\label{sec-etm-method-ot}
Optimal transport may be seen as a way to compute a global minimum cost alignment between sets of points given the cost of aligning individual pairs of points. Additionally, OT problems associate the sets with probability distributions where valid alignments satisfying capacity constraints specified over these distributions. Solutions to the OT problem result in sparse and soft alignments between the two sets when given meaningful pairwise costs from most scoring models, this makes them well suited to topic modeling while leveraging expressive LM/LLM models. 
 
 Specifically, we assume the source and target points to be document and label sets, $D$ and $L$, of size $n$ and $m$ respectively. We assume them to be distributed according to distributions $\mathbf{x}_D$ and $\mathbf{x}_{L}$, with pairwise costs $\mathbf{C}\in \mathbb{R}^{ n\times m}_+$. Since the distributions are unknown, we treat $\mathbf{x}_D$ and $\mathbf{x}_{L}$ to be uniform distributions. The solution to the OT problem is a soft assignment, the \emph{transport plan} $\mathbf{Q}^*$, which converts $\mathbf{x}_D$ into $\mathbf{x}_{L}$ by transporting probability mass from $\mathbf{x}_D$ to $\mathbf{x}_L$ while minimizing an aggregate cost $\mathcal{W}$, referred to as the Wasserstein Distance. Capacity constraints on $\mathbf{Q}$ result in a constrained linear optimization problem:
\begin{equation}
    {\mathcal{W}} = \underset{\mathbf{Q} \in \mathcal{S}}{\texttt{min}}\langle\mathbf{C},\mathbf{Q}\rangle
    \label{eq-ot-opt}
\end{equation}
 
We examine two kinds of capacity constraints on $\mathbf{Q}$ resulting in different feasible sets $\mathcal{S}$ -- a \emph{complete assignment} which assigns all source points to target points and a \emph{partial assignment} which does not require all source and target points to be assigned. This allows potentially noisy topic assignments to be ignored. Complete assignment ensures that the columns and rows of $\mathbf{Q}$ marginalize respectively to $\mathbf{x}_D$ and $\mathbf{x}_L$, resulting in $\mathcal{S}_{c}=\{\mathbf{Q} \in \mathbb{R}^{n\times  m}_{+}|\mathbf{Q}\mathbf{1}_{ m}=\mathbf{x}_{D}, \mathbf{Q}^{T}\mathbf{1}_{n}=\mathbf{x}_{L}\}$. Partial assignment requires only a fraction $p < 1$ of the source or target points to be assigned resulting in: $\mathcal{S}_{p}=\{\mathbf{Q} \in \mathbb{R}^{n\times  m}_{+}|\mathbf{Q}\mathbf{1}_{ m}
\leq\mathbf{x}_{D}, \mathbf{Q}^{T}\mathbf{1}_{n}\leq\mathbf{x}_{L}, \mathbf{1}^{T}\mathbf{Q}^{T}\mathbf{1}=p\}$. The mass $p$, to preserve in $\mathbf{Q}^*$ may be specified by analysts and captures the fraction of high quality topic assignments that can be made for $D$ and $L$.

In practice, for complete as well as partial assignments we solve an entropy regularized variant of Eq \eqref{eq-ot-opt}: $\mathcal{W} = \texttt{min}\langle\mathbf{C},\mathbf{Q}\rangle - 1/\lambda H(\mathbf{Q})$. While exact solutions to Eq \eqref{eq-ot-opt} require $O(n^3)$ computations, entropy regularization allows both problems to be solved using iterative methods with an empirical complexity of $O(n^2)$ and solvers capable of leveraging GPU computation. This allows scaling to larger datasets. To further speed up topic assignments for large $D$, we compute $\mathbf{Q}$ in batches of source points, with batches sized $<|D|$. To solve complete and partial batch assignment problems we leverage algorithms of \citet{cuturi2013sinkhorn} and \citet{benamou2015partialot} implemented in off the shelf solvers.\footnote{OT solvers: \url{https://pythonot.github.io}} Topic assignments are described further in \S\ref{sec-etm-hard-assignment} and Algorithm \ref{alg-edtm-topicassign}.
\begin{algorithm}[t]
\small
    \caption{Complete topic assignment in \etmmethod}
    \begin{algorithmic}[1]
        \State \textbf{Input}: $D$, $L$, $f_{\text{dist}}$
        \State $\mathbf{C} \gets f_{\text{dist}}(D,L)$ \Comment{Compute pairwise costs}
        \State $\mathbf{Q}^* \gets \text{zeros}(|D|, |L|)$ \Comment{Initialize plan with zeros}
        \For {epoch $e$ of $E$}
            \For {batch $b$ of $B$}
                \State $\mathbf{x}^b_D$, $\mathbf{x}_L \gets \text{uniform}(b), \text{uniform}(L)$
                \State $\mathbf{Q}^{*b} \gets \underset{\mathbf{Q}^b \in \mathcal{S}_c}{\texttt{argmin}}\langle\mathbf{C}^b,\mathbf{Q}^b\rangle - 1/\lambda H(\mathbf{Q}^b)$
                \State $\mathbf{Q}^*[b, :] \gets \mathbf{Q}^{*b}$ \Comment{Update plan for batch}
            \EndFor
        \EndFor
        \State $\mathbf{Q}^* \gets \mathbf{Q}^*/B\cdot E$ \Comment{Normalize plan}
        \State $\{l^*\}_{i=1}^{|D|} \gets \mathtt{argmax}_{l} \mathbf{Q}^*$
\end{algorithmic}
\label{alg-edtm-topicassign}
\end{algorithm}

\subsection{Document Topic Scoring}
\label{sec-etm-method-affinity}
To allow users to provide rich natural language topic targets $L$, we leverage LM/LLM models to compute document-topic costs $\textbf{C}[d,l]$ in Eq \eqref{eq-ot-opt}. Here, we leverage off-the-shelf BERT based bi-encoders and expressive T5-XL crossencoders trained for predicting query-document relevance in IR tasks \cite{nogueira2020document, lin2020pretrained}. With a bi-encoder we compute pairwise costs as L2 distances between document and label embeddings: $\text{L2}(f_{\text{BE}}(d), f_{\text{BE}}(l))$. From crossencoders, $f_{\text{CE}}$, we first obtain the probability of relevance $P_{\text{CE}}(\text{rel}=1|d,l)$. Then we obtain costs as: $1-{P_{\text{CE}}}/{max_{l}P_{\text{CE}}}$. The normalization per input $d$ ensures that costs remain comparable across all inputs with a minimum value of 0 across all documents. We note that while computation of costs with crossencoders is an expensive operation, recent work \cite{yadav2022anncur} leveraging matrix factorization to efficiently compute crossencoder based scores for large corpora promise a ready pathway toward scaling our approach to larger corpora -- we leave exploration of this to future work. We refer to bi-encoder and crossencoder variants as \etmmethodbe and \etmmethodce. We provide further implementation details in \S\ref{sec-etm-setup}.

\subsection{Hardening Topic Assignments}
\label{sec-etm-hard-assignment}
Here we outline the final procedure for obtaining hard topic assignments for documents from $\mathbf{Q}^*$. Both partial and complete topic assignments in $\mathbf{Q}^*$ may assign documents to a variable number of topics, however we only retain the top topic assignment per document for to make comparisons to baseline methods that make single topic assignments, representing the majority of recent baselines that we compare to. We leave exploration of multi-topic assignments to future work. Complete assignments are made as: $\mathtt{argmax}_{l} \mathbf{Q}^*$. For partial assignment, we first marginalize $\mathbf{Q}^*$ over the topic labels as: $\mathbf{x}^*_{D} = \sum_{l}\mathbf{Q}[\cdot, l]$. Then, we only make assignments for the fraction $p$ of documents which have the highest values in $\mathbf{x}^*_{D}$ (adding a step after Line 11 in Algorithm \ref{alg-edtm-topicassign}). Recall that $p$ represents the fraction of mass conserved in making partial assignments by the optimal transport solution of Eq \ref{eq-ot-opt}. In practice, points which don't receive assignments represent high-cost topic assignments which are unlikely to receive accurate assignments.

\section{Experiments}
\label{sec-etm-exp}
We experiment with \etmmethod in a variety of English datasets commonly used to evaluate topic-models. We compare \etmmethod against methods for few-shot classification leveraging LM/LLMs, clustering methods, and LDA based topic models. We evaluate \etmmethod using extrinsic metrics commonly used to evaluate clustering since we propose \etmmethod as a model for data exploration. Further, we evaluate \etmmethod in various interactive setups leveraging varying human interaction and noise. We outline our primary experimental setup next, and detail interaction setups in the respective sections.

\subsection{Experimental Setup}
\label{sec-etm-setup}
\begin{table*}[t]
\centering
\scalebox{0.8}{
\begin{tabular}{rlllll}
\toprule
Dataset    & $|D|$ & $|L|$ & $|d|$ & Label name and frequency \\
\midrule
Twitter    & $4373$ & $6$ & $28$ &\textcolor{MidnightBlue}{\texttt{pop culture}}, $1705$; \textcolor{MidnightBlue}{\texttt{sports and gaming}}, $1528$; \textcolor{MidnightBlue}{\texttt{daily life}}, $647$\\
Wiki       & $8024$ & $15$ & $2888$ & \textcolor{MidnightBlue}{\texttt{Media and drama}}, $1118$; \textcolor{MidnightBlue}{\texttt{Warfare}}, $1112$; \textcolor{MidnightBlue}{\texttt{Music}}, $1007$\\
Bills      & $15242$ & $21$ & $215$ & \textcolor{MidnightBlue}{\texttt{Health}}, $1755$; \textcolor{MidnightBlue}{\texttt{Public Lands}}, $1355$; \textcolor{MidnightBlue}{\texttt{Domestic Commerce}}, $1295$\\
Bookgenome & $9177$ & $226$ & $170$ & \textcolor{MidnightBlue}{\texttt{fiction}}, $548$; \textcolor{MidnightBlue}{\texttt{adventure}}, $352$; \textcolor{MidnightBlue}{\texttt{suspense}}, $301$\\
\bottomrule
\end{tabular}}
\caption{Summary of the datasets used for experiments: number of documents ($|D|$), number of labels ($|L|$), and average length of documents ($|d|$). We also present the top three labels and their frequencies, illustrating the topic names and label skews present in our datasets.}
\label{tab-etm-datasumm}
\end{table*}
\textbf{Datasets.} We use four datasets of short- to medium-length texts, each accompanied by a corresponding gold label. These datasets feature a diverse range of domains, including Wikipedia articles, Congressional Bills summaries, Twitter posts, and Goodreads book descriptions. As a result, the four datasets exhibit varying types and numbers of labels ($|L|$). Our datasets and their topic labels are summarized in Table \ref{tab-etm-datasumm}.

\ul{Twitter} contains short tweets from Sep 2019 to Aug 2021 paired with 1 of 6 high level labels assigned by crowd workers \cite{antypas2022twitter}.\footnote{HF Datasets: \href{https://huggingface.co/datasets/cardiffnlp/tweet_topic_single/viewer/tweet_topic_single/train_all}{\texttt{cardiffnlp/tweet\_topic\_single}}}

\ul{Wiki} contains Wikipedia articles designated to be ``Good articles'' based on adherence to editorial standards by Wikipedia editors. The articles are paired with 1 of 15 high-level and 114 finer-grained labels~\citep{merity2018regularizing}.

\ul{Bills} contains US congress bill summaries from January 2009 to January 2017 manually paired with 1 of 21 high-level and 45 finer-grained labels~\citep{adler2018congressional, hoyle2022neural}.

\ul{Bookgenome} contains book descriptions from Goodreads scored by crowd workers against 727 user-generated tags~\citep{kotkov2022bookgenome}. Of these we retain only the max scoring tag per book, and exclude tags which are used fewer than 5 times. This results in 226 tags which contain a mix of high level and finer grained tags, which serve as the gold labels in our evaluations of \etmmethod (\S\ref{sec-etm-interact-results}).

\textbf{Evaluation metrics.} We evaluate \etmmethod using extrinsic clustering evaluation metrics by comparing predicted clusters $\mathcal{C}'$, to gold clusters $\mathcal{C}$ induced by predicted and gold label assignments respectively. We report the set overlap based metric $P_1$~\cite{zhao2001criterion, amigo2009comparison} and the mutual information (MI), $I(\mathcal{C},\mathcal{C}')$ \cite{meila2007compareclusters}. $P_1$ is the harmonic mean of cluster purity and inverse purity, and is bounded between 0 and 1. While purity and inverse purity may be trivially maximized by over or under-segmenting $D$ into $1$ or $|D|$ clusters respectively, the harmonic mean trades these quantities off. Notably, $P_1$ captures the user experience in data exploration setups -- the coherence of clusters experienced by a user while accounting for over-segmentation. While $P_1$ represents a cluster level metric, MI represents a instance level metric -- telling us the reduction in uncertainty of the label for a point according to $\mathcal{C}$ if we know its labeling according to $\mathcal{C}'$.

In our evaluations we don't employ the Normalized Mutual Information, NMI: $I(\mathcal{C},\mathcal{C}')/(H(\mathcal{C})+H(\mathcal{C}'))$ commonly used in prior work since NMI results in higher metrics for imbalanced clusterings i.e clusterings with low entropy ($H(C)$ or $H(C')$), even if lower in mutual information $I(\mathcal{C},\mathcal{C}')$. Notably, all our datasets represent a realistic imbalanced labeling raising the chances of inflated NMI. Further, we don't rely on the Adjusted Rand Index (ARI) since it evaluates if pairs of points belong to the same cluster in $\mathcal{C}$ and $\mathcal{C}'$ -- while meaningful, pairwise relationships are less aligned with analyst workflows of cluster exploration where topic models are commonly employed. 

\textbf{Baselines.} We compare \etmmethod against various few-shot classification and clustering/topic modeling approaches. Clustering approaches range: \ul{LDA}: Represents a widely used approach to topic modeling representing documents as mixtures of latent topics, in turn represented as mixtures of the corpus vocabulary. Document-topic distributions are used for topic assignment. We use the MALLET implementation of LDA with Gibbs sampling \cite{mccallum2002mallet}. We set $|V|=15,000$, $\alpha=1.0$, $\beta=0.1$, and run LDA for 2,000 iterations with optimization at every 10 intervals. \ul{BertTopic}: Represents an approach to topic modeling implemented in the widely used \texttt{BERTTopic} package \cite{grootendorst2022bertopic}. Topic modeling is performed by embedding input texts with a pre-trained  BiEncoder\footnote{HF Model: \href{https://huggingface.co/sentence-transformers/all-MiniLM-L6-v2}{\texttt{all-MiniLM-L6-v2}}}, reducing dimensionality with UMAP, and clustering resulting embeddings using HDBSCAN. \ul{KMeans}: A standard clustering-based approach~\citep{macqueen1967some, lloyd1982least} to topic modeling, embedding inputs using the pretrained bi-encoder $f_{\text{BE}}$ used in \etmmethod then performing KMeans clustering. This has shown to be an effective approach to topic modeling \cite{thompson2020topic} -- notably however, we cluster document embeddings rather than contextualized word embeddings. \ul{TopicGPT}: A LLM based topic modeling approach, which involves prompting {GPT-4} to generate topics based on a small subset of $D$, and then using {GPT-3.5-Turbo} for assigning the generated topics to all documents in $D$ \cite{pham2023topicgpt}. For LDA, BertTopic, and KMeans we set the number of clusters to be $|L|$. The number of clusters in TopicGPT vary across datasets\footnote{$k=15$ for Twitter, $31$ for Wiki, $79$ for Bills, and $482$ for Bookgenome.} since we cannot fix the number of topics generated by {GPT-4}. Note here, that while we report the performance of TopicGPT we preclude extensive comparison to it given its use of two highly performant commercial LLMs, instead we treat it as an upper bound in performance of existing approaches.

\begin{table*}
\centering 
\scalebox{1}{
\begin{tabular}{rrcccccccc}
\toprule
&  \multicolumn{2}{c}{Bookgenome} & \multicolumn{2}{c}{Bills} & \multicolumn{2}{c}{Wiki} & \multicolumn{2}{c}{Twitter}\\
& \multicolumn{2}{c}{$|L|$: 226} & \multicolumn{2}{c}{$|L|$: 21} & \multicolumn{2}{c}{$|L|$: 15} & \multicolumn{2}{c}{$|L|$: 6}\\
\cmidrule(lr){2-3}\cmidrule(lr){4-5}\cmidrule(lr){6-7}\cmidrule(lr){8-9}
Method & P$_1$ & MI & P$_1$ & MI & P$_1$ & MI & P$_1$ & MI\\
\midrule
TopicGPT & \emph{0.18} & \emph{1.97} & \emph{0.57} & \emph{1.66} & \emph{0.73} & \emph{1.79} & \emph{0.75} & \emph{0.70}\\
LDA & {0.17} & 1.37 & \ul{0.56} & \ul{1.30} & \ul{0.73} & \ul{1.62} & 0.49	 & 0.29\\
BertTopic & 0.15 & 1.04 & 0.39 & 0.93 & 0.52 & 1.17 & 0.53 & 0.31\\
KMeans & 0.16 & {1.77} & 0.46 & 1.27 & 0.55 & 1.37 & \ul{0.55} & \ul{0.58}\\
GPT-3.5-Turbo & \ul{0.25} & \ul{1.92} & 0.51 & 1.21 & 0.71 & 1.59 & \ul{0.55} & 0.21\\
\midrule
NN$_\text{BE}$ & {\textbf{0.17}} & 1.74 & 0.52 & 1.19 & \textbf{0.57} & 1.10 & \textbf{0.64} & \textbf{0.57}\\
\etmmethodbe & {\textbf{0.17}} & {\textbf{1.84}} & \textbf{0.54} & \textbf{1.22} & \textbf{0.57} & \textbf{1.11} & 0.62 & 0.55\\
NN$_\text{CE}$ & {\textbf{0.20}} & 1.76 & \ul{\textbf{0.58}} & \ul{1.34} & 	0.61 & 1.25 & 0.71 & 0.68\\
\etmmethodce & {\textbf{0.20}} & {\textbf{1.77}} & \ul{\textbf{0.58}} & \ul{\textbf{1.35}} & \textbf{0.65} & \textbf{1.37} & \ul{\textbf{0.72}} & \ul{\textbf{0.70}}\\
\bottomrule
\end{tabular}}
\caption{Cluster quality with \etmmethod. \ul{Underline} for a baseline represents the best evaluation metric, and \ul{underline} for NN and \etmmethod~ indicate better or matched performance to the best baseline. \textbf{Bold} indicates the better performing model between NN and \etmmethod. We compare most closely to NN given that it represents the most similar approach to \etmmethod~ as we note in \S\ref{sec-etm-setup}.}
\label{tab-edtm-main}
\end{table*}
While clustering approaches remain agnostic to analyst provided label names, zero- and few-shot classification approaches leverage analyst provided label names, these baselines range: \ul{GPT3.5-Turbo}: This approach uses few-shot prompting to assign one of the $L$ labels to each input document with {GPT3.5-Turbo}. \ul{Nearest Neighbor (NN)}: This nearest neighbor model predicts the label most similar to the input text using a similarity metric identical to \etmmethod (\S\ref{sec-etm-method-affinity}), i.e labels are predicted as $\mathtt{argmin}_{l} \mathbf{C}$. We differentiate Bi-Encoder and crossencoder approaches as NN$_{\text{BE}}$ and NN$_{\text{CE}}$. Note that NN may seen as the greedy version of \etmmethod, making label assignments greedily for each $d\in D$ and may be seen as most similar to \etmmethod. For label assignment with TopicGPT and GPT-3.5-Turbo, we truncate the input document and our topic list if their combination exceed the LLM context length of 4,096 tokens.\footnote{We truncate either the topic list or the document if either component exceeds approximately half of the input context length (around 1,700 tokens).} Given the large label space in Bookgenome, truncation of the topic list is necessary. In such cases, we include only a top set of candidate labels, which are selected based on their cosine similarity with the input text as computed by a pretrained Bi-Encoder.\footnote{HF Model: \href{https://huggingface.co/sentence-transformers/all-MiniLM-L6-v2}{\texttt{all-MiniLM-L6-v2}}}

\textbf{Implementation Details.} For computing document-topic costs $\mathbf{C}$ in \etmmethod, for $f_{\text{BE}}$ we use a 110M parameter BERT-like Bi-Encoder pre-trained for dense retrieval on weakly supervised query-document pairs constructed from web-forums. For $f_{\text{CE}}$, we leverage the MonoT5 crossencoder based on T5-XL and trained for query-document relevance on the MS-MARCO dataset \cite{nogueira2020document}. Owing to the size of Bookgenome, we leverage a T5-Large crossencoder to keep experiments feasible.\footnote{HF Models; $f_{\text{BE}}$: \href{https://huggingface.co/sentence-transformers/multi-qa-mpnet-base-cos-v1}{\texttt{multi-qa-mpnet-base-cos-v1}}, $f_{\text{CE}}$: \href{https://huggingface.co/castorini/monot5-3b-msmarco-10k}{\texttt{monot5-3b-msmarco-10k}}} While, the average number of tokens in input texts across our datasets of Table \ref{tab-etm-datasumm} are $28$, $2888$, $215$, and $170$ -- we retain the first $450$ tokens to meet length limitations of LM/LLMs. Further, in inputing label text to $f_{\text{BE}}$ we format labels as questions, e.g.\ for Wiki, ``Is this a Wikipedia article about \texttt{LABEL}?'' to mimic the structure of training data for $f_{\text{BE}}$. Next, for computing complete assignments with optimal transport, we set the entropy regularizer $\lambda = 1$, compute $\mathbf{Q}$ in batches $b$ of size $500$ and averaged over 3 epochs. In experiments we distinguish bi-encoder and crossencoder versions as \etmmethodbe and \etmmethodce.

\subsection{Main Results}
\label{sec-etm-main-results}
Table \ref{tab-edtm-main} compares \etmmethod against baseline clustering as well as zero- and few-shot prediction approaches on 4 datasets varying in characteristics.

\textbf{Baseline performance.} We begin by examining the performance of baseline models. First, we note that LDA shows strong performance compared to other clustering models, KMeans and BertTopic. Next, LDA sees significantly lower performance on shorter Twitter texts and sees performance nearing that of TopicGPT on the significantly longer texts in Wiki -- this trend mirrors prior results of poor performance on short texts \cite{hong2010twittertm}. Next, we consider the performance of GPT-3.5-Turbo. First, we note that GPT-3.5-Turbo sees consistently lower performance compared to TopicGPT in 3 of 4 datasets -- indicating an inability of GPT-3.5-Turbo to make high quality assignments with analyst provided labels $L$. However, it sees stronger performance in Bookgenome. As we note in \S\ref{sec-etm-setup}, due to its limited context length, GPT-3.5-Turbo \emph{re-ranks} labels, this differs from the three other datasets -- relying on a first stage retrieval of labels using a bi-encoder followed by LLM based assignment and likely explains its strong performance. This may also represent a meaningful strategy for label supervised topic modeling with large $L$ that may be explored in future work.

\textbf{\etmmethod performance.} We begin by examining \etmmethod compared with NN. Here, note that \etmmethodbe matches or outperforms NN$_{\text{BE}}$ in three of four datasets and \etmmethodce matches or outperforms NN$_{\text{CE}}$ in all four datasets -- indicating the value of joint assignment of texts to labels over the greedy assignment of NN. Next, we note that use of a crossencoder improves upon the results of a bi-encoder in both NN and \etmmethod indicating the value of more expressive text similarity models. Finally, we compare the performance of \etmmethodbe and \etmmethodce with the best baseline approaches. Here we see that \etmmethod results in improvements compared to clustering approaches in Bookgenome, Bills, and Twitter. This may be attributed to the effective use of label name supervision absent in clustering approaches. Further, \etmmethod methods also outperform GPT-3.5-Turbo based assignment in Bills and Twitter data indicating their value in domains likely to be missing from LLM pretraining data. Finally, \etmmethodce also approaches the performance of TopicGPT in Bills, Bookgenome, and Twitter indicating its ability to induce high quality clusters at par with large scale LLMs while adhering to analyst provided topic labels.


\section{Interaction Experiments}
\label{sec-etm-interact-results}
In Tables \ref{tab-etm-seedwords}, \ref{tab-etm-seeddoc}, and \ref{tab-etm-partial} we present results in various interactive scenarios demonstrating respectively, the ability of \etmmethod to incorporate finer grained label descriptions, seed documents, and make high quality topic assignments in the presence of incomplete topic names. We make complete assignments for label descriptions and seed documents and partial assignments for incomplete topic names. For each interactive evaluation we first describe the experimental setup and follow with a discussion of experimental results.

\begin{table}[t]
\centering 
\scalebox{1}{
\begin{tabular}{rrcccc}
\toprule
&  \multicolumn{2}{c}{Bills} & \multicolumn{2}{c}{Wiki}\\
&  \multicolumn{2}{c}{$|L|$: 21} & \multicolumn{2}{c}{$|L|$: 15}\\
\cmidrule(lr){2-3}\cmidrule(lr){4-5}
Method & P$_1$ & MI & P$_1$ & MI\\
\midrule
GPT-3.5-Turbo & \ul{0.51} & 1.21 & \ul{0.71} & 1.59\\
SeededLDA & 0.48	& \ul{1.22} & 0.65 & \ul{1.71}\\
\midrule
NN$_\text{BE}$ & \ul{0.55}	& \ul{1.27} & 0.65 & 1.32\\
\etmmethodbe & \ul{\textbf{0.58}} & \ul{\textbf{1.31}} & \textbf{0.67} & \textbf{1.36}\\
NN$_\text{CE}$ & \ul{0.63} & \ul{1.44} & \ul{0.72} & 1.56\\
\etmmethodce & \ul{\textbf{0.65}} & \ul{\textbf{1.48}} & \ul{\textbf{0.74}} & \textbf{1.62}\\
\bottomrule
\end{tabular}}
\caption{Cluster quality with finer grained supervision provided per label in the form of seed words. \textbf{Bold} indicates better performance between NN and \etmmethod, and \ul{underline} indicates better performance compared to a baseline.}
\label{tab-etm-seedwords}
\end{table}
\subsection{Seed words as topic labels}
\label{sec-etm-interact-seedwords}
\textbf{Setup.} In this experiment we simulate a scenario where an analyst authors longer form topic descriptions instead of topic names alone. We limit experiments to the Bills and Wiki dataset, and use their finer grained topic labels to generate descriptions for each topic. We format these finer grained topics into a natural language description for  the target label set $L$, for example: ``Is this a wikipedia article about Media and drama or Television or Film or Actors?''. For NN and \etmmethod, these richer labels are used to compute document topic costs $\mathbf{C}$ that are used for topic assignment with bi-encoders or crossencoders. Here, we also compare to SeededLDA \cite{jagarlamudi2012incorporating}, an LDA topic model incorporating user provided seed words into induced topics.

\textbf{Results.} In Table \ref{tab-etm-seedwords}, first we note that both NN and \etmmethod outperform GPT-3.5-Turbo and SeededLDA in Bills, and NN and \etmmethod outperform baselines in Wiki with crossencoders. Further, comparing to Table \ref{tab-edtm-main}, we note that addition of label descriptions consistently improved performance for NN and \etmmethod. This indicates NN and \etmmethod's ability to incorporate rich natural language topic labels from analysts. Finally, we note that across Bills and Wiki, \etmmethod consistently outperforms NN with bi-encoder and crossencoder text similarities -- indicating joint assignments to benefit from improved similarities/cost estimates.

\subsection{Seed documents as topic labels}
\label{sec-etm-interact-seeddocs}
\begin{table}
\centering 
\scalebox{0.7}{
\begin{tabular}{rrcccccccc}
\toprule
&  \multicolumn{2}{c}{Bookgenome} & \multicolumn{2}{c}{Bills} & \multicolumn{2}{c}{Wiki} & \multicolumn{2}{c}{Twitter}\\
& \multicolumn{2}{c}{$|L|$: 226} & \multicolumn{2}{c}{$|L|$: 21} & \multicolumn{2}{c}{$|L|$: 15} & \multicolumn{2}{c}{$|L|$: 6}\\
\cmidrule(lr){2-3}\cmidrule(lr){4-5}\cmidrule(lr){6-7}\cmidrule(lr){8-9}
Method & P$_1$ & MI & P$_1$ & MI & P$_1$ & MI & P$_1$ & MI\\
\midrule
NN$_\text{BE}$ & \textbf{0.23} & 1.91 & \textbf{0.54} & 1.25 & 0.53 & 0.97 & \textbf{0.64} & \textbf{0.46}\\
\etmmethodbe & 0.19 & \textbf{1.94} & \textbf{0.54} & \textbf{1.26} & \textbf{0.56} & \textbf{1.08} & 0.56 & 0.41\\

\bottomrule
\end{tabular}}
\caption{Cluster quality with topic labels represented with averaged embeddings of high precision seed documents retrieved using the target label with a retrieval model.} 
\label{tab-etm-seeddoc}
\end{table}
\textbf{Setup.} Here, we simulate a scenario where analysts use topic names in $L$ to perform a search over the corpus $D$, verifies their correctness, and uses the verified sample documents as topic targets. This setup also mirrors a common scenario where seed documents serve as queries for corpus exploration \cite{wang2021journalisticsource}. Here, we compare NN and \etmmethod~ alone given that few-shot classification with GPT-3.5-Turbo runs into context length limitations in using document examples for our datasets. Further, we only experiment with bi-encoder variants given that crossencoders remain limited by context length limitations for larger number of seed documents. For NN$_\text{BE}$ and \etmmethodbe~ we compute costs $\mathbf{C}$ between documents $d$ and topic labels $l$, using the top five verified retrievals per label as: $\text{L2}[f_{\text{BE}}(d), \mathtt{mean}_{k=1\dots 5}f_{\text{BE}}(d_l^k)]$. We choose $k=5$ to represent a reasonable effort to verify label-document correctness by an analyst.

\textbf{Results.} In Table \ref{tab-etm-seeddoc}, we note that \etmmethodbe matches or outperforms NN$_\text{BE}$ in Bills, Wiki, and Bookgenome (MI) indicating its ability to incorporate seed document supervision. We also note, comparing to Table \ref{tab-edtm-main} that while NN$_\text{BE}$ and \etmmethodbe sees improvement from using seed documents in Bookgenome and Bills, they see drops in performance in datasets with fewer labels, Wiki and Twitter. This follows from the finer grained labels of Bookgenome and Bills being better represented using high precision seed documents. However, in Wiki and Twitter, seed documents are only likely to represent certain aspects of the higher level labels. The resulting lower quality document-topic similarities result in especially degraded assignments in \etmmethodbe on Twitter. However, in the presence of finer grained labels in the remaining three datasets \etmmethod results in high quality topic assignments.

\subsection{Partial assignment of topics}
\label{sec-etm-interact-partial}
\begin{table}
\centering 
\scalebox{0.7}{
\begin{tabular}{rrcccccccc}
\toprule
&  \multicolumn{2}{c}{Bookgenome} & \multicolumn{2}{c}{Bills} & \multicolumn{2}{c}{Wiki} & \multicolumn{2}{c}{Twitter}\\
& \multicolumn{2}{c}{$|L|$: 226} & \multicolumn{2}{c}{$|L|$: 21} & \multicolumn{2}{c}{$|L|$: 15} & \multicolumn{2}{c}{$|L|$: 6}\\
\cmidrule(lr){2-3}\cmidrule(lr){4-5}\cmidrule(lr){6-7}\cmidrule(lr){8-9}
Method & P$_1$ & MI & P$_1$ & MI & P$_1$ & MI & P$_1$ & MI\\
\midrule
NN$_\text{BE}$ & \textbf{0.18} & \textbf{1.76} & \textbf{0.51} & \textbf{1.19} & \textbf{0.58} & \textbf{1.13} & \textbf{0.65} & \textbf{0.54}\\
\etmmethodbe & 0.17 & \textbf{1.76} & \textbf{0.51} & 1.17 & 0.56 & 1.09 & 0.64 & 0.52\\
NN$_\text{CE}$ & \textbf{0.21} & \textbf{1.79} & \textbf{0.55} & 1.27 & 0.62 & 1.29 & {0.71} & \textbf{0.65}\\
\etmmethodce & 0.20 & 1.76 & \textbf{0.55} & \textbf{1.31} & \textbf{0.64} & \textbf{1.33} & \textbf{0.73} & \textbf{0.65}\\
\bottomrule
\end{tabular}}
\caption{Cluster quality with a topic label omitted from the label set $L$, following which NN and \etmmethod~ make partial assignments i.e witholding predictions for certain inputs. This setup simulating a scenario where analysts may not list all topic labels due to insufficient knowledge of a corpus. Reported numbers are averaged over metrics obtained from excluding 3 high frequency topic labels one at a time.}
\label{tab-etm-partial}
\end{table}
\textbf{Setup.} To demonstrate the value of making partial assignments we simulate a scenario where a label is missing from the target label set $L$, e.g.\ due to an analyst having insufficient knowledge of the corpus topics. This creates a scenario where a performant model should not make topic assignments for some inputs. Specifically, we exclude one label, $l_e$ from $L$ selected at random from the most frequent 5 labels and make a partial assignment with NN or \etmmethod. We repeat this procedure with 3 different $l_e$ labels and report averaged performance over 3 different clusterings. In each case, $p$ is set to the proportion of documents with $l_e$ in the gold labeling. For making partial assignments we follow the procedure outlined in \S\ref{sec-etm-hard-assignment}. In computing evaluation metrics we exclude input texts which don't receive a cluster assignment from NN and \etmmethod.

\textbf{Results.} In Table \ref{tab-etm-partial} we note that with bi-encoder costs, NN slightly outperforms or matches the performance of \etmmethod. However, with crossencoder costs, \etmmethodce sees stronger performance than NN. Note here that in the presence of missing labels in $L$, both NN and \etmmethod could make alternative topic assignments for documents which could have received $l_e$ while leaving a fraction $p$ of documents unassigned. For example, when \texttt{Language and literature} topic name is excluded from $L$ in Wiki, \etmmethodce most frequently makes assignments to: \texttt{Media and drama}, \texttt{History}, \texttt{Philosophy and religion}, and others. Manual examination revealed these to often be reasonable. However, since we only consider single topic assignments in our evaluations evaluating alternative topic assignments or multi-topic assignments, more generally, remains future work. Nevertheless, these results indicate the ability of \etmmethod to make high quality topic assignments despite receiving incomplete topic sets $L$ indicating an ability to handle the errors likely in interacting with users.

\section{Related Work}
\label{sec-etm-related}
We begin by discussing prior work on interactive and supervised topic modeling. Then we discus prior work leveraging optimal transport and large language models for topic modeling.

\textbf{Interactive topic modeling.} Topic modeling has focused on learning human interpretable topics given only a corpus of documents. A vast body of work has explored probabilistic generative models for topic modeling, representing topics as distributions over word types and documents as mixtures of latent topics \cite{jbg2017tmbook, lund2019fine}. 
We first examine the line of work which has sought to incorporate supervision from users into these models. Supervision from users has been of three broad types: (1) document labels (e.g.\ sentiment) and metadata (eg.\ dates) paired with each document, (2) user specified seed-words available at the corpus level, and (3) through constraints (e.g.\ ``must-link'') over latent topics. While work on (1) and (2) have not historically been considered interactive topic modeling, we consider them as such.

Incorporation of document level labels and metadata has been explored through generation of labels conditioned on latent topics \cite[STM]{mcauliffe2007supervised}, conditioning document generation on observed metadata \cite{card2018neural}, or as priors influencing document-topic distributions\cite[DMR]{mimno2008topic}. While these approaches aim to influence latent topics with labels/metadata, \citet[LabeledLDA]{ramage2009labeledlda} sought to ensure one-to-one correspondence between latent topics and document multi-labels, learning a word-label assignment. In ensuring a correspondence between topics and labels LabeledLDA bears resemblance to \etmmethod. However, in contrast with this line of work, \etmmethod does not assume knowledge of documents paired with labels or metadata.

In assuming corpus level topic labels our work resembles prior work that seeks to incorporate user provided seed words for topics. Here, work of \cite{jagarlamudi2012incorporating, churchill2022guidedtm} incorporates topic-seed words into the generation of document words with a mixture of seeded and latent topic-word distributions or through modified sampling schemes. On the other hand, \citet{bahareh2022keyetm} incorporate seed information into an embedded topic model by regularizing the topic-word matrix. Relatedly, \citet{akash2022coordinated} incorporate weak supervision into topic-word and document-topic matrices from a LabeledLDA and zero-shot classification model respectively. \etmmethod extends topic models based on seed words to more verbose forms of topics supervision such as descriptions and documents through its use of LM/LLM similarity functions, while also showing partial assignments to result in accurate topic assignments in the presence of noise in user provided topics.

Finally, a third line of work prior work has explored interaction with topic models through ``must-link'' and ``cannot-link'' constraints available a-priori \cite{andrzejewski2009incorporating} or supplied interactively once a model of topics has been learned \citet{hu2011interactive} with tree structured priors for word-topic distributions. Interaction through such constraints provides a complementary form of interaction than ours -- we leave exploration of such constraints into \etmmethod to future work.

While a large body of work has explored incorporation of various forms of interaction into generative topic models such interactions have also been explored for topic modeling in other frameworks. The Anchor algorithm for topic modeling based on non-negative matrix factorization \cite{lund2019fine} was extended to incorporate document metadata as well as seed word supervision from users \cite{nguyen2015supervisedanchors, lund2017tandemanchor}. \citet{pacheco2021drail} introduce a probabilistic programming framework for relational learning and demonstrate its use for interactive topic modeling via first order logic rules \cite{pacheco2023interactive}. In contrast with these approaches \etmmethod offers a more natural form of interaction with topic models through long form natural language interactions by leveraging LM/LLMs for modeling text similarities. Work of \citet{meng2020discriminative} learns a discriminative model for retrieving words representative of a topic given only a topic name, bootstrapped from pre-trained word-embeddings. While mirroring \etmmethod in its use of topic names it remains limited to exploring greedy assignments similar to our nearest neighbor (NN) baseline. Finally, while a large body of work, including \etmmethod, has explored use of supervision to guide latent topics toward user provided topics, \citet{thompson2018authorless} highlight the value of biasing topics away from user labels and discovering more novel topical structure \cite{ramage2011partiallda} -- we leave exploration of such considerations in \etmmethod to future work.

\textbf{LLM topic models.} The advent of highly performant LM/LLMs has lead recent work to explore use of these models for topic modeling and text clustering. The dominant line of work here has explored unsupervised topic modeling. These approaches commonly consist of two stages, topic generation with an LLM followed by a topic assignment to the generated topics. A considerable design space has been explored for both stages. \citet[TopicGPT]{pham2023topicgpt} leverages GPT-4 for topic generation and GPT-3.5-Turbo for topic assignment, we include TopicGPT in our experiments. Other two stage approaches have also explored iterative generation of topic taxonomies rather than flat topic lists \cite{lam2024concept, wan2024tnt}. For topic assignment, \citet{lam2024concept} explore multi-choice assignment with LLMs while \citet{wan2024tnt} train small classifiers on LLM assignments to speed up test time topic assignment. Notably, these approaches omit joint assignment of documents to topics as in \etmmethod. In this regard, \citet[GoalEx]{wang2023goalex} bears resemblance to \etmmethod, leveraging integer linear programs for global document-topic assignment following LLM based topic generation and initial greedy assignment. Differing from GoalEx, \etmmethod's use of optimal transport for assignment allows richer document-topic costs to be used and fractional assignments to be made rather than single topic  assignments of GoalEx -- however, both forms of joint assignment may be valuable in various applications. Also similar in its use of joint assignment to topic names is the work of \citet{fei2022beyondprompting} -- leveraging greedy classification with pretrained embedding models followed by joint document-topic assignment with Gaussian Mixture Models. However use of GMM's precludes use of black-box similarity functions such as crossencoders possible to use in \etmmethod.

Besides leveraging the zero/few shot classification ability of LLMs for topic modeling as in \etmmethod, a smaller body of work has also leveraged LLMs for refining text clustering. \citet{viswanathan2023large} explore using LLMs to augment input texts to clustering algorithms, generate oracle constraints for constrained clustering methods, and iteratively re-assign low confidence cluster assignments. On the other hand, \citet{zhang2023clusterllm} explore actively training text similarity models based on LLM similarities and determining cluster granularities with LLMs. Leveraging LLMs for actively refining clustering presents an underexplored line of future work.

\textbf{Optimal transport for topic models.} A small body of work has leveraged optimal transport for topic modeling. \citet{zhao2020neural} learn encoders for transforming bag of word document representations to topic distributions by minimizing OT distances between the two distributions -- building on embedded topic models \cite{dieng2020topic}. On the other hand, \citet{huynh2020otlda} model topic modeling as a dictionary learning problem, representing documents as mixtures of latent topics which are learned by minimizing the Wasserstain distance to bag of word document representations. While these approaches must learn latent topics, we explore optimal transport for supervised or interactive topic modeling. While this simplifies the problem to an assignment problem, it introduces complexities such as dealing with noise in labels -- to the best of our knowledege no prior works have explored optimal transport in this setup and \etmmethod represents initial work in this space. 

Finally, \citet{mysore2023lace} leverage the label-name interactions of \etmmethod for developing a controllable recommendation model that represents historical user documents using a model similar to \etmmethodbe -- in contrast, this work explores various other forms of interaction, partial assignments, and crossencoder similarities for interactive topic modeling. However, future work may explore other downstream applications while leveraging a \etmmethod to represent a collection of texts.
\section{Conclusion}
\label{sec-etm-conclusion}
In this paper in introduce \etmmethod, a label supervised topic model. \etmmethod leverages optimal transport based assignment algorithms to make globally coherent topic assignments for documents based on pre-trained LM/LLM based document-topic affinities. The proposed method results in high quality topics compared to a range of baseline methods based on pretrained LM/LLMs, LDA topic models, and clustering methods. \etmmethod is also shown to incorporate numerous different forms of interaction from analysts while remaining robust to noise in analysts' input.

\section{Acknowledgments}
This work was supported in part by awards IIS-2202506 and IIS-2312949 from the National Science Foundation (NSF), the Center for Intelligent Information Retrieval, Lowe’s, IBM Research AI through the AI Horizons Network, and Chan Zuckerberg Initiative under the project Scientific Knowledge Base Construction. Any opinions, findings and conclusions or recommendations expressed in this material are those of the authors and do not necessarily reflect those of the sponsor.

\bibliography{editable_topic_models}
\bibliographystyle{acl_natbib}



\end{document}